\documentclass[letterpaper]{article} 
\pdfoutput=1
\usepackage{aaai23}  
\usepackage{times}  
\usepackage{helvet}  
\usepackage{courier}  
\usepackage[hyphens]{url}  
\usepackage{graphicx} 
\usepackage{natbib}  
\usepackage{caption} 
\usepackage{amsmath,amsfonts}
\usepackage{algorithmic}
\usepackage{algorithm}
\usepackage{array}
\usepackage{textcomp}
\usepackage{stfloats}
\usepackage{url}
\usepackage{verbatim}
\usepackage{graphicx}
\usepackage{color}
\usepackage{multirow}
\usepackage{arydshln}
\usepackage{bbding}
\usepackage{wrapfig}
\usepackage{caption}
\usepackage{booktabs}
\usepackage{colortbl}
\usepackage{newfloat}
\usepackage{listings}

\usepackage{newfloat}
\usepackage{listings}
\DeclareCaptionStyle{ruled}{labelfont=normalfont,labelsep=colon,strut=off} 
\floatstyle{ruled}
\newfloat{listing}{tb}{lst}{}
\floatname{listing}{Listing}
\makeatletter

\newcommand{\Rmnum}[1]{\expandafter\@slowromancap\romannumeral #1@}
\makeatother
\title{Self Correspondence Distillation for End-to-End Weakly-Supervised Semantic Segmentation}
\author{
Rongtao Xu\textsuperscript{\rm 1,3,}\equalcontrib,
Changwei Wang\textsuperscript{\rm 1,3,}\equalcontrib,
Jiaxi Sun\textsuperscript{\rm 1,3},
\\Shibiao Xu\textsuperscript{\rm 2,}\thanks{Shibiao Xu and Weiliang Meng are the corresponding authors.},
Weiliang Meng\textsuperscript{\rm 1,3,\dag},
Xiaopeng Zhang\textsuperscript{\rm 1,3}
}
\affiliations{
    \textsuperscript{\rm 1}NLPR, Institute of Automation, Chinese Academy of Sciences\\
    \textsuperscript{\rm 2}School of Artificial Intelligence, Beijing University of Posts and Telecommunications\\
    \textsuperscript{\rm 3}School of Artificial Intelligence, University of Chinese Academy of Sciences\\
    shibiaoxu@bupt.edu.cn, \{xurongtao2019, wangchangwei2019, sunjiaxi2020, weiliang.meng, xiaopeng.zhang\}@ia.ac.cn
    }

\usepackage{bibentry}

\begin{document}

\maketitle

\begin{abstract}
Efficiently training accurate deep models for weakly supervised semantic segmentation (WSSS) with image-level labels is challenging and important. Recently, end-to-end WSSS methods have become the focus of research due to their high training efficiency. However, current methods suffer from insufficient extraction of comprehensive semantic information, resulting in low-quality pseudo-labels and sub-optimal solutions for end-to-end WSSS. To this end, we propose a simple and novel Self Correspondence Distillation (SCD) method to refine pseudo-labels without introducing external supervision. Our SCD enables the network to utilize feature correspondence derived from itself as a distillation target, which can enhance the network's feature learning process by complementing semantic information.
In addition, to further improve the segmentation accuracy, we design a Variation-aware Refine Module to enhance the local consistency of pseudo-labels by computing pixel-level variation. Finally, we present an efficient end-to-end Transformer-based framework (TSCD) via SCD and Variation-aware Refine Module for the accurate WSSS task. Extensive experiments on the PASCAL VOC 2012 and MS COCO 2014 datasets demonstrate that our method significantly outperforms other state-of-the-art methods. 
Our code is available at {https://github.com/Rongtao-Xu/RepresentationLearning/tree/main/SCD-AAAI2023}.




\end{abstract}

\section{Introduction}
Weakly Supervised Semantic Segmentation (WSSS) aims to use weak/cheap labels to alleviate the reliance on time-consuming pixel-level annotations~\cite{zhang2021adaptive}. The most challenging scenario for WSSS methods is to use only image-level labels which we focus on, as it has no real localization information, and usually relies on multi-stage solutions to generate high-quality pseudo-labels.

\begin{figure}[t]
\begin{center}
  \includegraphics[width=1 \linewidth]{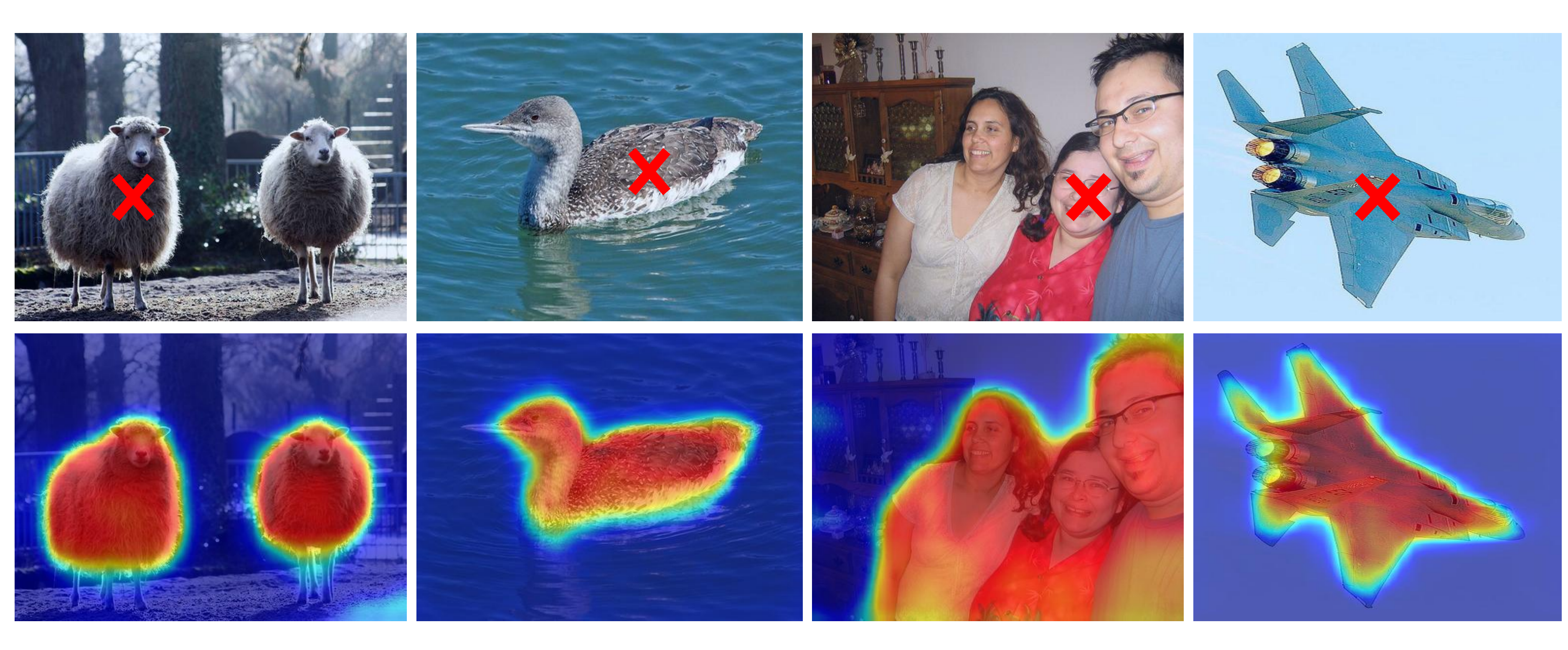}
\end{center}
   \caption{Visualization examples of CAMs generated by our TSCD. The corresponding categories are drawn on the original image with red crosses.
}
\label{fig:intro}
\end{figure}

Multi-stage methods~\cite{hou2018self_e,lee2019ficklenet,wu2021embedded_s} first train a classification model and then train a semantic segmentation network with refined pseudo-labels, which are generated from class activation maps~\cite{zhou2016learning} (CAMs). Due to the complex pipeline of multi-stage methods and the need to train multiple models, recently some end-to-end methods~\cite{araslanov2020single,zhang2020reliability,zhang2021adaptive,ru2022learning} have been proposed to speed up the training process. However, these methods rely heavily on the pseudo-labels generated by CAM as supervision information and still show limited performance in obtaining the comprehensiveness of semantic information and segmentation accuracy.
The recent work~\cite{caron2021dino} reveals that there is corresponding information about image semantic segmentation in the dense features of self-supervised transformers. We find that these dense features can be class activation maps, and further construct novel CAM feature correspondence to learn semantic correspondence information.
In this paper, we propose a simple and novel method named Self Correspondence Distillation (SCD), which can enhance the network feature learning process by complementing semantic information to refine CAM without external supervision.
Our SCD allows the network to utilize the CAM feature correspondence derived from itself as the distillation target for segmentation prediction features. This  mechanism can help the network obtain comprehensive image semantic information and improve accuracy.

To further refine the initial pseudo-labels obtained by CAM, we design a Variation-aware Refine Module (VARM), which introduces the idea of image noise reduction, and updates pseudo-labels by calculating image pixel-level variation and image local information.
Based on SCD and VARM, we provide an end-to-end Transformer-based framework with Self Correspondence Distillation (TSCD) for WSSS.
We also visualize the CAM generated by our TSCD in Figure~\ref{fig:intro}, and the results show that our TSCD can generate fine-grained CAMs by capturing comprehensive image semantic information.

In summary, our main contributions are as follows:
\begin{itemize}
\item We propose a novel Self Correspondence Distillation method, namely SCD, to enhance the feature learning of WSSS networks. According to our knowledge, this is the first attempt to use the network's own CAM feature correspondence as the distillation target.
\item We design the Variation-aware Refine Module (VARM) to address the local inconsistency of pseudo-labels. Our VARM refines the CAM by computing image pixel-level variation and employing pixel-adaptive convolution.
\item We provide an end-to-end Transformer-based framework (TSCD) with Self Correspondence Distillation and VARM for weakly supervised semantic segmentation. Our TSCD achieves state-of-the-art segmentation performance on PASCAL VOC 2012 and MS COCO 2014.
\end{itemize}

\section{Related Work}
\subsection{Weakly-Supervised Semantic Segmentation}
Recently, deep learning has been developed rapidly~\cite{wang2022cndesc,wang2022mtldesc,xu2022domaindesc,xu2023rssformer,zhao2021attention}, especially in image segmentation~\cite{xu2021dc,xu2022instance,wang2022softgan,wang2021cganet,wang2022net,Dong_CVPR2020}.
Weakly-Supervised Semantic Segmentation (WSSS) methods are mainly divided into two types: Multi-stage Methods and End-to-End Methods, both of which rely on class activation maps~\cite{zhou2016learning} to generate pseudo labels. For Multi-stage Methods, the erasure methods~\cite{hou2018self_e} make the classifier focus on the complete object region by erasing the discriminative region. To obtain multiple attribute maps of target objects, Lee et al.~\cite{lee2019ficklenet} utilize dilated convolutions to accumulate activation regions, while \cite{wu2021embedded_s} explore semantic similarities and differences across multiple input images. Several recent approaches attempt to explain and improve CAM generation, such as~\cite{zhang2020causal} from the perspective of causal relationships between images and labels, and~\cite{lee2021reducing} from the perspective of information bottleneck theory. Due to the rough boundary of the obtained initial localization map, refinement techniques such as CRF and IRN~\cite{ahn2019weakly} are applied for subsequent refinements.

While most WSSS methods are based on multi-stage, recent End-to-End Methods have also achieved comparable performance to multi-stage methods. For End-to-End Methods, training a well-performing end-to-end model is challenging due to the extremely limited supervision brought by the rough initial CAM. To address this problem, 
many studies have focused on improving pseudo-segmentation labels obtained from CAM maps.
Araslanov et al.~\cite{araslanov2020single} proposed 1Stage which achieved great progress on the WSSS task via pixel-adaptive mask refinement, while Zhang et al.~\cite{zhang2020reliability} used CRF on CAM to generate fine labels as supervision.
Besides, Zhang et al.~\cite{zhang2021adaptive} introduce feature-to-prototype alignment loss and adaptive affinity field, and Ru et al.~\cite{ru2022learning} learn affinity from attention to refine CAM.
Different from the above methods, we refine the CAM by learning the feature correspondence of the network's own CAMs.

\subsection{Knowledge Distillation and Self-Supervised Feature Learning}
Knowledge distillation~\cite{hinton2015distilling_d} is a model compression method by transferring knowledge from large models to small models. Usually the student network and the teacher network share the same capacity, and the student network imitates the intermediate output of the teacher network. Recent studies~\cite{hou2019learning_d} focus on attention distillation, where the student network learns attention maps obtained from the teacher network. Hou et al.~\cite{hou2019learning_d} proposed SAD that can learn attention knowledge without a teacher network. Different from~\cite{hou2019learning_d}, we exploit feature correspondence to obtain comprehensive image semantic information without teacher network, inspired by self-supervised feature learning.

The goal of self-supervised feature learning is to learn meaningful visual features without human annotation, while contrastive learning over multiple augmentations is used to address this problem~\cite{chen2020big}. Recently, Pinheiro et al.~\cite{o2020unsupervised} generate spatially dense feature maps to compare local pixel features. \cite{hamilton2022unsupervised} treats pretrained self-supervised features as input. Different from the above methods, our TSCD refines the CAM without external supervision, by taking the CAM feature correspondence as the distillation target, and our novel idea is that the CAM feature correspondence may be extracted for self-learning.

\section{Proposed Method}
\subsection{Preliminary}
The Class Activation Map (CAM) was first proposed in~\cite{zhou2016learning} to enable classifiers learn object localization. Given a feature map $f\in \mathbb{R}^{H\times W\times D}$ extracted by CNN and the class $c$, we apply global average pooling and fully connected layer operations to compute class scores:
\begin{equation}
\begin{aligned}
y_{c} = \frac{1}{HW}  \sum_{k}^{K} w_{c,k} \sum_{i}^{} f_{k,i}
\end{aligned}
\end{equation}
where $w_{c,k}$ represents the parameters of the fully connected layer. CAM is achieved by weighting the contribution of each channel of the feature map to a specific category of the classification score. Formally, the class activation map $m_{c}$ of the class $c$ is:
\begin{equation}
\begin{aligned}
m_{c} = Relu(\sum_{k}^{K} w_{c,k} f_{k,:} ).
\end{aligned}
\end{equation}
In this paper, we adopt the class activation map as the initial pseudo-label.

\subsection{Self Correspondence Distillation}
\label{sec:scd}
Our goal is to perform Self Correspondence Distillation to refine the CAM of the original image. Our Self Correspondence Distillation does not require any additional labels or external supervision, while it can assist the network to obtain comprehensive image semantic information. Recent advances in self-supervised feature learning have shown that dense features are semantically related~\cite{collins2018deep,zhou2016learning}. In this paper, we verify that these dense feature maps can be class activation maps, and further construct CAM feature correspondence for feature learning.

\begin{figure*}[htbp]
\begin{center}
  \includegraphics[width=0.9 \linewidth]{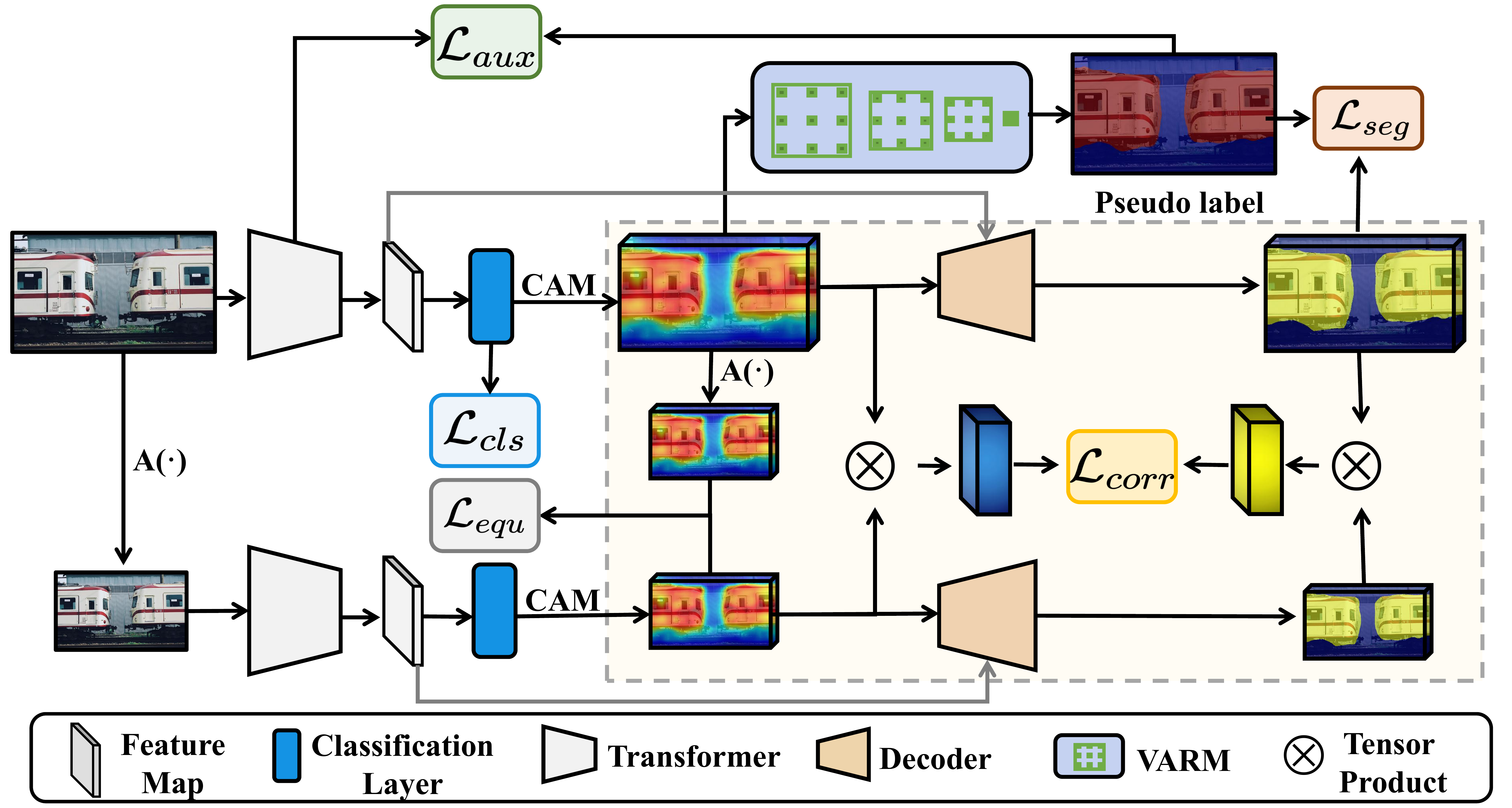}
\end{center}
   \caption{Illustration of the proposed end-to-end framework TSCD. Our TSCD uses the Transformer backbone as the encoder and adopts CAM~\cite{zhou2016learning} to generate initial pseudo-labels. The generated initial pseudo-labels are then refined with our SCD and VARM. The yellow area indicates the application of our SCD. The optimization of the network consists of self correspondence distillation loss, equivariant regularization loss, classification
   loss, and auxiliary loss, etc. $A(\cdot)$ means affine transformation.}
\label{fig:overview}
\end{figure*}

\subsubsection{CAM Feature Correspondence:}
Formally, we focus on the correlation volume between CAMs. Given two CAMs $m_{1}\in \mathbb{R}^{H_{1}\times W_{1}\times C}$ and $m_{2}\in \mathbb{R}^{H_{2}\times W_{2}\times C}$, with $H_{1},H_{2}$ for height, $W_{1},W_{2}$ for width, and $C$ for category, we define the CAM feature correspondence as:

\begin{equation}
\begin{aligned}
\mathcal{M}_{h_{1}w_{1}h_{2}w_{2}} = \sum_{c}^{} \frac{m_{1}\cdot  m_{2}}{\left | m_{1} \right |  \left | m_{2} \right |}.
\end{aligned}
\end{equation}

Specifically, given an image $I\in \mathbb{R}^{H\times W\times D}$ and an affine transformation $A(\cdot)$, we employ the feature maps extracted by the encoder to generate CAMs. Taking CAM $m_{1}$ as an example, we use $E: \mathbb{R}^{H\times W\times D} \to \mathbb{R}^{H_{1}\times W_{1}\times C}
$ to denote the process of CAM $m_{1}$ generation from the extracted feature maps. Therefore, $m_{1}$ and $m_{2}$ can be expressed as:

\begin{equation}
\begin{aligned}
\begin{cases}m_{1} = E(I)
 \\
m_{2} = E(A(I)) 
\end{cases}
\end{aligned}
\end{equation}

Then the CAM feature correspondence can be further expressed as:

\begin{equation}
\begin{aligned}
\mathcal{M}_{h_{1}w_{1}h_{2}w_{2}} = \sum_{c}^{} \frac{E(I) \cdot E(A(I))}{\left | E(I) \right |  \left | E(A(I)) \right |} 
\end{aligned}
\end{equation}
whose entries represent the cosine similarity between the feature at $m_{1}$'s position $(h_{1},w_{1})$ and $m_{2}$'s position $(h_{2},w_{2})$.
As shown in Figure~\ref{fig:overview}, the encoder and decoder share weights, and the segmentation prediction maps are denoted as $s_{1}\in \mathbb{R}^{H_{1}\times W_{1}\times C}$ and $s_{2}\in \mathbb{R}^{H_{2}\times W_{2}\times C}$ for images $I$ and $A(I)$, respectively. Then the segmentation feature correspondence is defined as:

\begin{equation}
\begin{aligned}
\mathcal{S}_{h_{1}w_{1}h_{2}w_{2}} = \sum_{c}^{} \frac{s_{1}\cdot  s_{2}}{\left | s_{1} \right |  \left | s_{2} \right |} .
\end{aligned}
\end{equation}

\subsubsection{SCD for Training:}
\label{sec:scdloss}
The intuition behind our Self Correspondence Distillation (SCD) is that segmentation feature correspondence can extract useful semantic information from the CAM feature correspondence to refine CAMs in a self-learning manner. Inspired by the self-supervised feature learning, we consider aligning the segmentation feature correspondence with the network's own CAM feature correspondence. The loss function is designed to push the corresponding CAMs together to enhance semantic relatedness, when there is a significant correlation between two segmentation predictions.  We implement the SCD loss function by simple element-wise multiplication of the segmentation feature corresponding $\mathcal{S}_{h_{1}w_{1}h_{2}w_{2}}$ and the CAM feature corresponding $\mathcal{M}_{h_{1}w_{1}h_{2}w_{2}}$:

\begin{equation}
\begin{aligned}
\mathcal{L}_{SCD} = - \sum_{h_{1}w_{1}h_{2}w_{2}}^{}\mathcal{M}_{h_{1}w_{1}h_{2}w_{2}} \cdot max(\mathcal{S}_{h_{1}w_{1}h_{2}w_{2}},0)
.
\end{aligned}
\end{equation}
where $max(\cdot,0)$ means the zero clamping. In practice, in order to ensure the inference efficiency, we adopt a random sampling strategy to train our SCD loss function, and the number of samples is $n$. If the size of the segmentation prediction map is different from the size of the corresponding CAM, then bilinear upsampling is applied to the segmentation prediction map.

\subsection{Variation-Aware Refine Module}
\label{sec:VARM}
The initial pseudo-labels obtained by CAM are usually locally inconsistent and have rough boundaries. Many multi-stage methods employ CRF to further refine the initial pseudo-labels, which reduces the training efficiency. For end-to-end methods,  Araslano et al.~\cite{araslanov2020single} utilize pixel-adaptive convolution~\cite{su2019pixel} to extract local image information for local consistency, while Ru et al.~\cite{ru2022learning} 
further combine spatial information to build a refinement module. Different from~\cite{araslanov2020single,ru2022learning}, we design the Variation-aware Refine Module, which introduces the idea of image noise reduction to overcome local inconsistencies.

Specifically, for positions $(i,j)$ and $(k,l)$ in image $I$, we first compute the image pixel-level variation:

\begin{equation}
\begin{aligned}
\mathcal{V}_{ij,kl} = (x_{i,j-1,kl} - x_{i,j,kl} )^{2} + (x_{i+1,j,kl} - x_{i,j,kl} )^{2}.
\end{aligned}
\end{equation}

Next we calculate the local information term $k^{rgb}$:
\begin{equation}
\begin{aligned}
k_{ij,kl}^{rgb} = \frac{-(\alpha \left | I_{i,j} - I_{k,l} \right |  )^{2} }{\sigma_{ij}  ^{2}}
\end{aligned}
\end{equation}
where $\sigma_{ij}$ represents the standard deviation and $\alpha$ represents the smoothing weight.
To enhance the local consistency of pseudo-labels, for pixels with large variation in the image, we calculate the correction kernel $k_{ij,kl}$ to avoid some abruptly deformed values:

\setlength\belowdisplayskip{1pt}
\begin{small} 
\begin{equation}
\begin{aligned}
k_{ij,kl} = \frac{exp(k_{ij,kl}^{rgb})}{\sum_{(x,y)\in N(i,j)}^{} exp(k_{ij,kl}^{rgb})} -  \beta  \frac{exp(\mathcal{V}_{ij,kl})}{\sum_{(x,y)\in N(i,j)}^{} exp(\mathcal{V}_{ij,kl})}
\end{aligned}
\end{equation}
\end{small} 
where $N (i, j)$ is the set of neighbors of $(i, j)$, obtained using dilated convolution as defined in~\cite{araslanov2020single}. We adopt an iterative update strategy to update the pixel label (CAM) $\mathcal{P}^{i,j,c}$:
\begin{equation}
\begin{aligned}
\mathcal{P}_{t}^{i,j,c} = \sum_{(k,l) \in N(i,j)}^{} k_{ij,kl} \mathcal{P}_{t-1}^{k,l,c}.
\end{aligned}
\end{equation}

Our Variation-aware Refine Module enhances the local consistency of the initial pseudo-labels by perceiving pixel-level variation, while ensuring high training efficiency.

\subsection{Transformer-Based Framework with Self Correspondence Distillation}
\label{sec:loss}
As shown in Figure~\ref{fig:overview}, our Transformer-based framework with Self Correspondence Distillation (TSCD) consists of a transformer backbone, SCD, VARM, equivariant regularization loss, classification loss, auxiliary loss and segmentation loss. Next, each loss function and total loss are introduced separately.

\textbf{Equivariant Regularization Loss:} Equivariant constraints have been shown to narrow the supervision gap between weak supervision and full supervision~\cite{du2021weakly,wang2020self}. We construct an equivariant constraint using an equivariant regularization loss:

\begin{equation}
\begin{aligned}
\mathcal{L}_{equ} = \left \| A(m_{1}) - m_{2}  \right \| _{1} = \left \| A(E(I)) - E(A(I))  \right \| _{1}.
\end{aligned}
\end{equation}

\textbf{Classification Loss:} As shown in Figure~\ref{fig:overview}, the aggregated feature maps are fed into the classification layer to compute prediction vector $p$ for the image-level classification. We adopt the multi-label soft-margin loss as the classification loss for network training. For the total number of classes $C$, the classification loss is defined as:

\begin{equation}
\begin{aligned}
\mathcal{L}_{cls} = \frac{1}{C} \sum_{c=1}^{C} (l^{c}log(\frac{1}{1+e^{-p_{c}}})+(1-l^{c})log(\frac{e^{-p_{c}}}{1+e^{-p_{c}}}))
\end{aligned}
\end{equation}
where $l$ is the ground truth for class labels.

\textbf{Auxiliary Loss:} To further enhance the performance of our network, we adopt the affinity loss of~\cite{ru2022learning} as our auxiliary loss in Figure~\ref{fig:overview}.
Affinity loss is beneficial to the self-attention learning of the encoder and helps the network focus on complete object regions. Different from~\cite{ru2022learning}, we directly use the attention maps $(\mathcal{A}_{1},\mathcal{A}_{2})$ output by the last two layers of the encoder to calculate the auxiliary loss. Formally, the auxiliary loss is expressed as:

\begin{equation}
\begin{aligned}
\mathcal{L}_{aux} = \frac{1}{N^{-} }  \sum_{R^{-}}^{} (1- \frac{1}{1+e^{-concat(\mathcal{A}_{1},\mathcal{A}_{2})} } ) +
\\ \frac{1}{N^{+} }  \sum_{R^{+}}^{} (\frac{1}{1+e^{-concat(\mathcal{A}_{1},\mathcal{A}_{2})} } )
\end{aligned}
\end{equation}
where concat means that attention maps are concatenated together,
$R^{+}$ and $R^{-}$ denote the set of positive and negative samples in the pseudo-affinity labels generated by refining pseudo-labels, while $N^{+}$ and $N^{-}$ denote the number of $R^{+}$ and $R^{-}$ respectively.

\textbf{Total Loss:}
We adopt the cross-entropy loss as the segmentation loss $L_{seg}$. Besides, to enhance the local consistency of segmentation results, we also adopt the commonly used regularization loss $L_{reg}$~\cite{tang2018_r,zhang2021adaptive}.
The total loss is defined as:

\begin{equation}
\begin{aligned}
\mathcal{L} = \lambda _{1}(\mathcal{L}_{corr} + \mathcal{L}_{seg}  + \mathcal{L}_{equ} + \mathcal{L}_{aux}) + \lambda _{2}\mathcal{L}_{reg} + \lambda_{3}\mathcal{L}_{cls}
\end{aligned}
\end{equation}

where $\lambda_{1}$,$\lambda_{2}$ and $\lambda_{3}$ are empirically set to $0.1$, $0.01$ and $1$ respectively.

\begin{figure}[t]
\begin{center}
  \includegraphics[width=1 \linewidth]{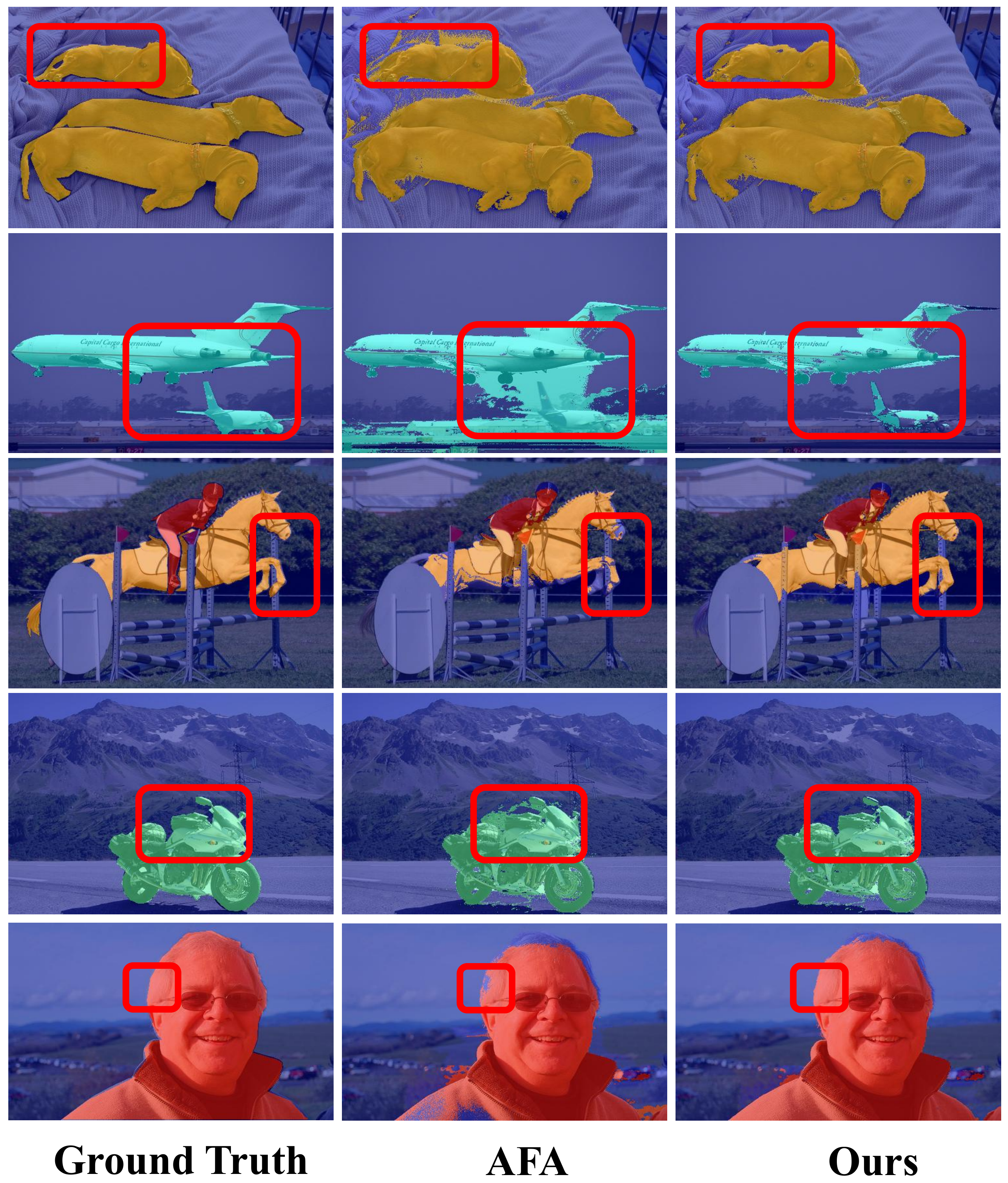}
\end{center}
   \caption{Examples of segmentation results of AFA~\cite{ru2022learning} and our method on PASCAL VOC validation images. The red box shows the difference.
}
\label{fig:voc}
\end{figure}

\begin{table}[t]
\centering
\renewcommand\arraystretch{1} {
\resizebox{\linewidth}{!}{
        \begin{tabular}{lclcc} 
\toprule
Method & \textit{Sup} & {Backbone} & \textit{val} & \textit{test} \\ 
\hline
\multicolumn{5}{l}{{\textbf{\textit{Fully-supervised methods}}}} \\
DeepLab & \multirow{3}{*}{\textit{F}} & R101 & 77.6 & 79.7 \\
Segformer &  & MiT-B1 & 78.7  & - \\
\midrule
\multicolumn{5}{l}{{\textbf{\textit{Multi-Stage~weakly-supervised methods}}}} \\
MCIS$_{ECCV’2020}$ & \multirow{3}{*}{\textit{I }+ \textit{S}} & R101 & 66.2  & 66.9 \\
AuxSegNet$_{ICCV’2021}$ &  & WR38 & 69.0 & 68.6 \\
EPS$_{CVPR’2021}$ &  & R101 & \textbf{70.9} & \textbf{70.8} \\ 
\midrule
SEAM$_{CVPR’2020}$ & \multirow{6}{*}{\textit{I}} & WR38 & 64.5 & 65.7 \\
SC-CAM$_{CVPR’2020}$ &  & R101 & 66.1 & 65.9 \\
CDA$_{ICCV’2021}$ &  & WR38 & 66.1 & 66.8 \\
AdvCAM$_{CVPR’2021}$ &  & R101 & 68.1 & 68.0 \\
CPN$_{ICCV’2021}$ &  & R101 & 67.8 & 68.5 \\
RIB$_{NeurIPS’2021}$ &  & R101 & \textbf{68.3} & \textbf{68.6} \\ 
\midrule
\multicolumn{5}{l}{{\textbf{\textit{End-to-End weakly-supervised methods}}}} \\
EM$_{ICCV’2015}$ & \multirow{9}{*}{\textit{I~}} & VGG16 & 38.2 & 39.6 \\
MIL$_{CVPR’2015}$ &  & - & 42.0 & 40.6 \\
CRF-RNN$_{CVPR’2017}$ &  & VGG16 & 52.8 & 53.7 \\
RRM$_{AAAI’2020}$ &  & WR38 & 62.6 & 62.9 \\
RRM+$_{AAAI’2020}$ &  & MiT-B1 & 63.5 & - \\
1Stage$_{CVPR’2020}$ &  & WR38 & 62.7 & 64.3 \\
AA$\&$LR$_{ACM MM’2021}$ &  & WR38 & 63.9 & 64.8 \\
AFA$_{CVPR’2022}$ &  & MiT-B1 & 63.8 & - \\
AFA+ CRF$_{CVPR’2022}$ &  & MiT-B1 & 66.0 & 66.3 \\ 
\cdashline{1-1}\cdashline{2-5}
\textbf{TSCD (Ours)} & \multirow{2}{*}{\textit{I~}} & MiT-B1 & 65.0 & 65.2 \\
\textbf{TSCD + CRF (Ours)} &  & MiT-B1 & \textbf{67.3 }& \textbf{67.5} \\
\bottomrule
\end{tabular}}}
\caption{Semantic segmentation comparisons of PASCAL VOC 2012 dataset. \textit{F}: full supervision; \textit{S}: saliency maps; \textit{I}: image-level labels.}
\label{tab:voc}
\end{table}

\section{Experiments}
\label{sec:exp}
\subsection{Experimental Settings}

\textbf{Dataset and evaluation metric:}
We evaluate our method on the commonly used PASCAL VOC 2012~\cite{everingham2010pascal} dataset and MS COCO 2014~\cite{lin2014microsoft} dataset. The PASCAL VOC 2012 dataset consists of training, validation and test sets with a total of 21 semantic classes. Usually the SBD dataset~\cite{hariharan2011semantic} is used for the augmentation of the PASCAL VOC 2012 dataset, and the augmented dataset includes 10,582 images for training, 1,449 images for validation, and 1,464 images for testing.
The MS COCO 2014 dataset~\cite{lin2014microsoft} contains 81 classes with training set and validation set,  each containing 82,081 and 40,137 images respectively.
Note that we only use image-level labels for annotation. We adopt the mean Intersection-Over-Union (mIoU) as the evaluation metric.

\textbf{Reproducibility:}
We implement our method using the PyTorch framework. For the training phase, we employ the AdamW optimizer with an initial learning rate set to $6 \times 10^{-5}$ and a weight decay factor 0.01. We use simple data augmentation strategies such as random rescaling, random horizontal flipping, and random cropping. We set the batch size to 8 and the crop size to $512\times 512$ . For VARM, we set $(\alpha,\beta)$ to (4, 0.01). Besides, we follow~\cite{ru2022learning} and set the dilation rate of the dilated convolution to $[1, 2, 4, 8, 12, 24]$. The number of samples $n$ for SCD loss is set to 40.
For pseudo-label generation, we follow~\cite{ru2022learning} and set the two background thresholds to 0.55 and 0.35, respectively.
For the experiments on PASCAL VOC 2012, the total number of iterations is set to 20,000, with 2,000 iterations warmed up for the classification branch. For the experiments of MS COCO 2014, the total number of iterations is set to 80,000, with 5,000 iterations of warm-up.

\textbf{Network Configuration:} 
Considering the training efficiency, our transformer backbone adopts the recent Segformer~\cite{xie2021segformer}.
For the encoder, we adopt the Mix Transformer (MiT)~\cite{xie2021segformer} and initialize the parameters with ImageNet-1k pretrained weights. Then we use the MLP decoder header~\cite{xie2021segformer} as the decoder to predict the refined pseudo-labels.

\subsection{Comparisons with State-of-the-Arts}
\textbf{PASCAL VOC 2012 dataset:}
For a fair comparison, we follow the evaluation code of~\cite{ru2022learning}.
We report mIoU values of state-of-the-art semantic segmentation methods and our method's mIoU values on VOC 2012 validation and test images in Table~\ref{tab:voc}.
It is observed that our TSCD outperforms all the other state-of-the-art end-to-end methods in VOC 2012.
These end-to-end methods include 1Stage~\cite{araslanov2020single} and AFA~\cite{ru2022learning}, etc.
R101 means using ResNet101~\cite{he2016deep} as the backbone, and WR38 means using WideResNet38~\cite{wu2019wider}. Our method also achieves comparable performance to recent multi-stage methods, including SEAM~\cite{wang2020self}, SC-CAM~\cite{chang2020weakly} and CDA~\cite{su2021context}, etc.
Compared to RRM~\cite{zhang2020reliability} and AFA~\cite{ru2022learning} using the same backbone, our method improves mIoU on the val set by 5\% and 2\%, respectively. Our TSCD even achieves 85.5\% of its fully supervised counterpart, Segformer~\cite{xie2021segformer}. This validates the effectiveness of our proposed Self Correspondence Distillation method (SCD) and Variation-aware Refine Module (VARM).

\begin{figure}[t]
\begin{center}
  \includegraphics[width=1 \linewidth]{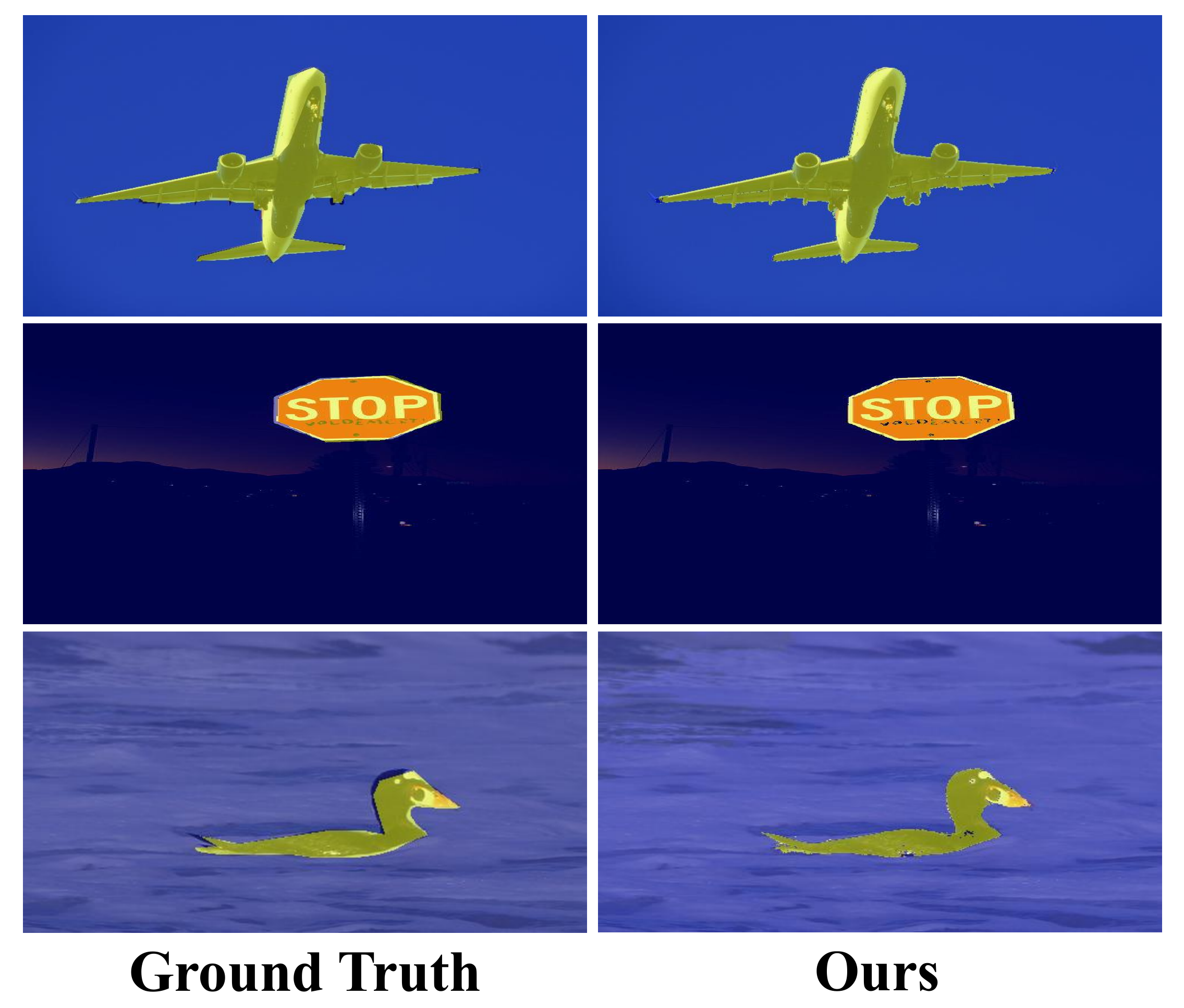}
\end{center}
   \caption{Examples of segmentation results of our method on COCO val images.
}
\label{fig:coco}
\end{figure}

\begin{table}[t]
\centering
\renewcommand\arraystretch{1} {
\resizebox{\linewidth}{!}{
        \begin{tabular}{lcll}
\toprule
{Method} & \textit{Sup} & Backbone & \textit{mIoU(\%)} \\ 
\midrule 
\multicolumn{4}{l}{\textbf{{\textit{Multi-Stage weakly-supervised methods}}}} \\
AuxSegNet$_{ICCV’2021}$ & \multirow{2}{*}{\textit{I }+ \textit{S}} & WR38 & 33.9 \\
EPS$_{CVPR’2021}$ &  & R101 & \textbf{35.7} \\ 
\midrule
SEAM$_{CVPR’2020}$ & \multirow{5}{*}{\textit{I~}} & WR38 & 31.9 \\
CONTA$_{NeurIPS’2020}$ &  & WR38 & 32.8 \\
CDA$_{ICCV’2021}$ &  & WR38 & 31.7 \\
CGNet$_{ICCV’2021}$ &  & WR38 & 36.4 \\
RIB$_{NeurIPS’2021}$ &  & R101 & \textbf{43.8} \\ 
\hline
\multicolumn{4}{l}{{\textbf{\textit{End-to-End weakly-supervised methods}}}} \\
AFA$_{CVPR’2022}$ & \multirow{2}{*}{\textit{I~}} & MiT-B1 & 38.0 \\
AFA + CRF$_{CVPR’2022}$\textbf{~} &  & MiT-B1 & 38.9 \\ \cdashline{1-4}
\textbf{\textbf{TSCD (Ours)}} & \multirow{2}{*}{\textit{I~}}  & MiT-B1 & 39.2 \\
\textbf{\textbf{TSCD + CRF (Ours)}} &  & MiT-B1 & \textbf{40.1} \\
\bottomrule
\end{tabular}}}
\caption{Semantic segmentation comparisons of MS COCO 2014 validation images.}
\label{tab:coco}
\end{table}

\textbf{MS COCO 2014 dataset:}
The mIoU values for the semantic segmentation on the MS COCO 2014 dataset are presented in Table~\ref{tab:coco}.
We report the semantic segmentation performance of multi-stage methods (EPS~\cite{lee2021railroad}, AuxSegNet~\cite{xu2021leveraging}, CDA~\cite{su2021context}, CGNet~\cite{kweon2021unlocking} and RIB~\cite{lee2021reducing}) and end-to-end methods AFA~\cite{ru2022learning}.
We can see that our TSCD significantly outperforms these state-of-the-art end-to-end methods and multi-stage methods except RIB~\cite{lee2021reducing}, reaching mIoU of 40.1\%. Compared with AFA using the same backbone, the mIoU of our method is improved by 3\%.

\begin{table}
\centering
\resizebox{\linewidth}{!}{
\begin{tabular}{cccccc} 
\toprule 
VARM & SCD & $\mathcal{L}_{aux}$ & {$\mathcal{L}_{euq}$} & {Method} & \textit{val} \\ 
\cmidrule{1-6}
 &  &  &  & Baseline & 47.2 \\ 
\midrule
\CheckmarkBold &  &  &  & \multirow{4}{*}{TSCD (Ours)} & 56.8 \\
\CheckmarkBold & \CheckmarkBold &  &  &  & {63.5} \\
\CheckmarkBold & \CheckmarkBold & \CheckmarkBold &  &  & {64.2} \\
\CheckmarkBold & \CheckmarkBold & \CheckmarkBold & \CheckmarkBold &  & \textbf{{65.0}} \\
\bottomrule
\end{tabular}}
\caption{Ablation studies on our TSCD framework. Here \CheckmarkBold indicates that this component is applied.} 
\label{tab:tscd}
\end{table}

\subsection{Ablation Studies}
We conduct various ablation studies on the PASCAL VOC 2012 dataset.

\textbf{Ablation studies of our TSCD framework:}
First, we verify the effectiveness of each part of our TSCD. Table~\ref{tab:tscd} reports the quantitative results of ablation analysis on the PASCAL VOC 2012 validation set. Our transformer-based baseline model achieves an mIoU of 47.2\%. Compared to the baseline, our VARM and SCD further significantly improve mIoU by 20\% and 34\%, respectively. Using the auxiliary loss $\mathcal{L}_{aux}$ and 
equivariant regularization loss $\mathcal{L}_{equ}$ brings the proposed framework to 64.2\% mIoU and 65.0\% mIoU, respectively.

\begin{table}[hb]
	\centering
\renewcommand\arraystretch{1} {
\begin{tabular}{lll} 
\toprule 
{Method} & {Backbone} & \textit{val} \\
\midrule
R101-Base & R101 & 53.5 \\
WR38-Base & WR38 & 56.2 \\
R101-SCD & R101 & 56.5 \\
WR38-SCD & WR38 & 59.1 \\
{TSCD $w/o$ SCD} & MiT-B1 & 63.9 \\ 
\hdashline
{\textbf{TSCD (Ours)}} & {MiT-B1} & \textbf{65.0} \\
\bottomrule
\end{tabular}}
\caption{Ablation studies on our SCD.}
\label{tab:scd}
\end{table}

\begin{figure*}[t]
\begin{center}
  \includegraphics[width=1 \linewidth]{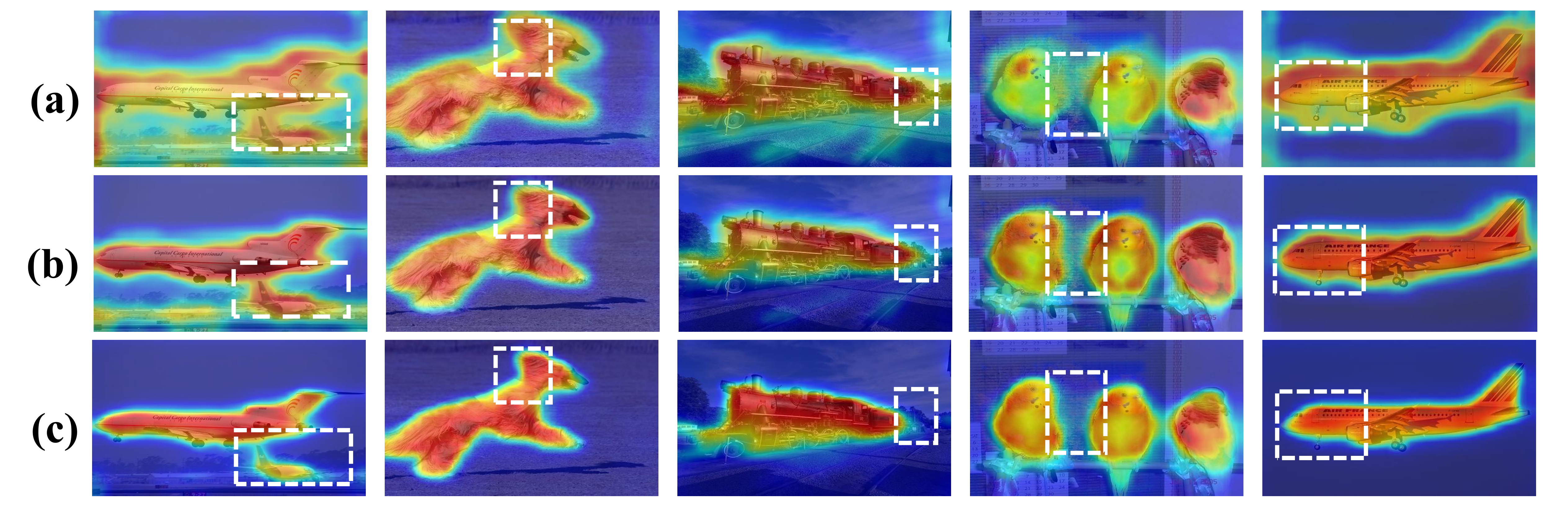}
\end{center}
   \caption{Visualization examples of CAMs. From top to bottom: (a) CAMs generated by our baseline; (b) CAMs after applying our VARM; (c) CAMs after applying our SCD.
}

\label{fig:tscd}
\end{figure*}

The experiments in Table~\ref{tab:tscd} validate the effectiveness of individual modules in our framework.
We also visualized the CAM of the baseline and the CAMs after applying our VARM and SCD respectively. As shown in Figure~\ref{fig:tscd}, the results show that VARM can further refine CAM, and SCD can enable the network to capture complete image semantic information and improve the quality of pseudo-labels.

\textbf{Ablation studies on SCD:}
To verify the generalization and effectiveness of our SCD, we also apply it to different mainstream backbones for weakly supervised semantic segmentation. 
We named the ResNet101~\cite{he2016deep} backbone we implemented for the WSSS task as R101-Base, and the application of SCD on the R101-Base as R101-SCD.
Implementations based on the WideResNet38~\cite{wu2019wider} backbone are named WR38-Base and WR38-SCD,  similar as ResNet101.
The quantitative results in Table~\ref{tab:scd} show that both R101-SCD and WR38-SCD improve the network performance compared to the original backbone network, reaching mIoU of 56.5\% and 59.1\% on the VOC val, respectively.
After removing the SCD from our TSCD, the network's ability to obtain comprehensive semantic information is weakened, resulting in a considerable decrease in the network's performance.

\begin{figure*}[t]
    \centering
	\begin{minipage}{0.47\linewidth}
		\centering
		\setlength{\abovecaptionskip}{0.28cm}
		\includegraphics[width=\linewidth]{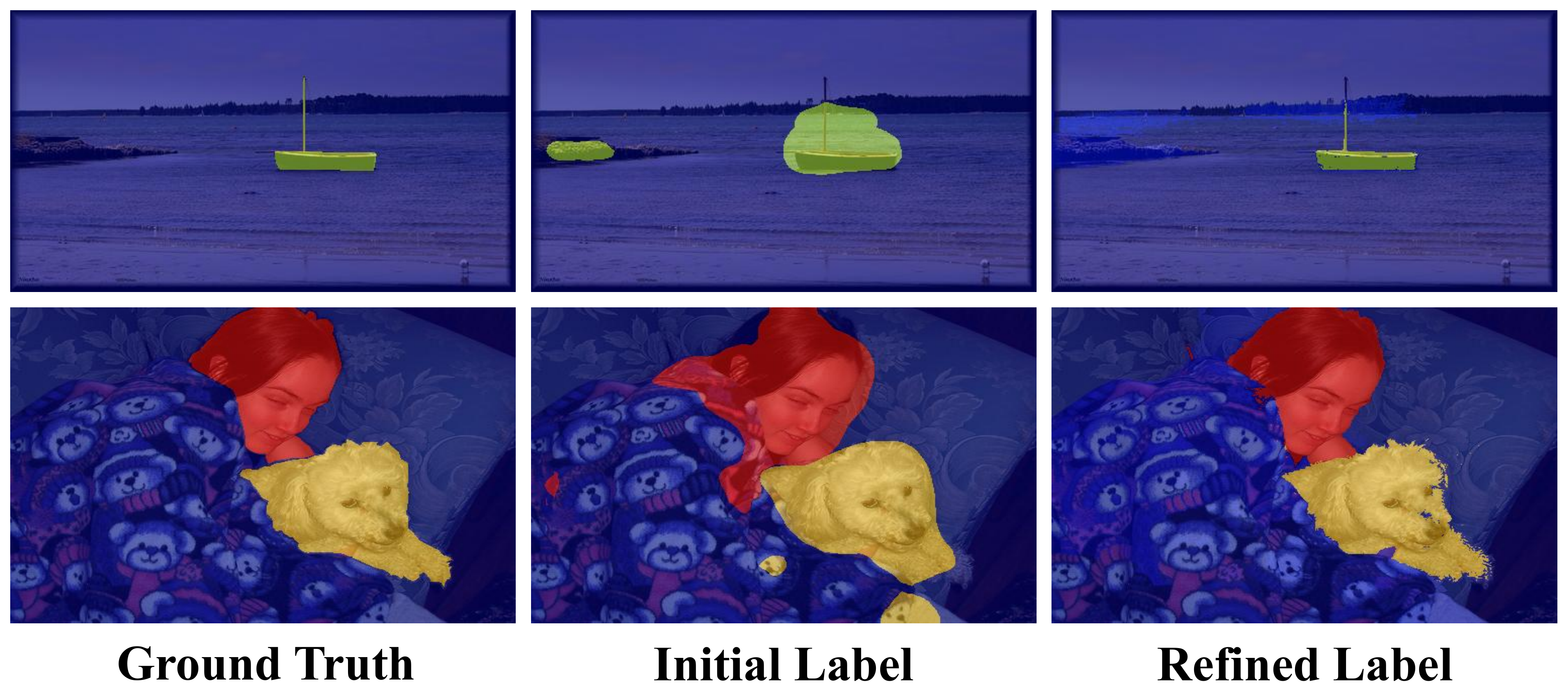}
		\caption{Visualization of pseudo-label refinement via VARM. Pseudo-labels are generated using CAM and our baseline.}
		\label{fig:varm}
	\end{minipage}
	\hfill
	\begin{minipage}{0.48\linewidth}
		\centering

		\renewcommand\arraystretch{0.8} {
		\resizebox{\linewidth}{!}{
        \begin{tabular}{cccc} 
\toprule 
$k^{rgb}$ & {$\mathcal{V}$} & {Method} & \textit{val} \\ 
\midrule
 &  & {Baseline} & 47.2 \\
\midrule
\CheckmarkBold &  & PAMR~\cite{araslanov2020single} & 54.3 \\
\CheckmarkBold &  & PAR~\cite{ru2022learning} & 54.9 \\ 
\midrule
\CheckmarkBold &  & \multirow{2}{*}{VARM} & 55.0 \\
\CheckmarkBold & \CheckmarkBold &  & \multicolumn{1}{l}{\textbf{56.8}} \\
\bottomrule
\end{tabular}}}
        
        \captionof{table}{Ablation studies on VARM and comparison with other refinement methods.}
        \label{tab:varm}
	    \end{minipage}
\end{figure*}

\textbf{Ablation studies on VARM:}
Our VARM aims to overcome local inconsistencies by perceiving pixel-level variation, thereby refining the initial pseudo-labels. Table~\ref{tab:varm} shows the impact of each component in VARM on the segmentation results, and the comparison with other refinement methods such as PAMR~\cite{araslanov2020single} and PAR~\cite{ru2022learning}. We use the CAM generated by the transformer-based 
baseline as the initial pseudo-label. The quantitative results in Table~\ref{tab:varm} show that our VARM can effectively improve the segmentation accuracy. Compared to baseline, our VARM improves mIoU by 20\% ($47.2 \to 56.8$). Our VARM, PAMR~\cite{araslanov2020single} and PAR~\cite{ru2022learning} are all based on dilated pixel-adaptive convolutions, and our method outperforms PAMR~\cite{araslanov2020single} and PAR~\cite{ru2022learning}, which validates the effectiveness of VARM.
In addition, Figure~\ref{fig:varm} shows that our VARM can enhance the local consistency of pseudo-labels and make the segmentation boundaries clearer, which makes the refined pseudo-labels closer to the ground truth.
See supplementary material for more details for our SCD and VARM.

\section{Conclusion}
We address the challenges of WSSS using image-level class labels in obtaining comprehensive semantic information and high segmentation accuracy, and propose a simple and novel Self Correspondence Distillation method (SCD) to refine CAM without additional labels. Our SCD uses the network's own CAM feature correspondence as the distillation target to enhance the feature learning process of the network. To further refine the initial pseudo-labels, we design the Variation-aware Refine Module (VARM), which enhances the local consistency of pseudo-labels by perceptual pixel-level variation. Based on SCD and VARM, an end-to-end Transformer-based framework (TSCD) for WSSS is provided. Extensive experiments demonstrate the effectiveness of our method, achieving the state-of-the-art performance on the PASCAL VOC 2012 and MS COCO 2014 datasets.

\section{Acknowledgements}
This work was supported in part by the National Natural Science Foundation of China (Nos. $U21A20515$, $62271074$, $61971418$, U$2003109$, $62171321$, $62071157$ and $62162044$) and in part by the Open Research Fund of Key Laboratory of Space Utilization, Chinese Academy of Sciences (No. LSU-KFJJ-2021-05), Open Research Projects of Zhejiang Lab (No. 2021KE0AB07), and this work was supported
by the Open Projects Program of National Laboratory of Pattern Recognition.

\bibliography{aaai23}

\end{document}